\documentclass[11pt,a4paper]{article}
\usepackage{times,latexsym}
\usepackage{url}
\usepackage[T1]{fontenc}

\usepackage{graphicx}
\usepackage{booktabs}
\usepackage{multicol}
\usepackage{multirow}
\usepackage{amssymb}% http://ctan.org/pkg/amssymb
\usepackage{mathtools}
\usepackage{pifont}% http://ctan.org/pkg/pifont
\usepackage{bbold}
\usepackage{caption}
\usepackage{subcaption}
\usepackage[normalem]{ulem}

\usepackage[acceptedWithA]{tacl2021v1}
\usepackage{xspace,mfirstuc,tabulary}

\newif\iftaclinstructions
\taclinstructionsfalse % AUTHORS: do NOT set this to true
\iftaclinstructions
\renewcommand{\confidential}{}
\renewcommand{\anonsubtext}{(No author info supplied here, for consistency with
TACL-submission anonymization requirements)}
\newcommand{\instr}
\fi

\iftaclpubformat % this "if" is set by the choice of options

\else

\fi

\title{Explainable Abuse Detection as Intent Classification and Slot Filling}

\author{
    Agostina Calabrese
\and
  Bj{\"o}rn Ross
  \and
  Mirella Lapata
  \\
  Institute for Language, Cognition and Computation
  \\
  School of Informatics, University of Edinburgh
  \\
  10 Crichton Street, Edinburgh EH8 9AB, United Kingdom
  \\
  \texttt{\{a.calabrese@,b.ross@,mlap@inf.\}ed.ac.uk}
}

\date{}

\begin{document}
\maketitle

%Introduce area
%Introduce problem
%Existing solutions
%Your method
%A summary of how the solution was evaluated and the result of the evaluation
\begin{abstract}
%\textit{Warning: this paper discusses and contains hate speech.}
To proactively offer social media users a safe online experience, there is a need for systems that can detect  harmful posts and promptly alert platform moderators. In order to guarantee the enforcement of a consistent policy,  moderators are provided with detailed guidelines. In contrast, most state-of-the-art models learn what \textit{abuse} is from labelled examples and as a result base their predictions on spurious cues, such as the presence of group identifiers, which can be unreliable. In this work we  introduce the concept of policy-aware abuse detection, abandoning the unrealistic expectation that systems can reliably learn which phenomena constitute abuse from inspecting the data alone.
% Instead, we equip them with a machine-friendly representation of the policy to be enforced. We collect the first dataset for explainable policy-aware abuse detection, consisting of 3535 English sentences. We show how architectures for intent classification and slot filling can be used to address the task in a%n explainable and
% more robust fashion. We make data and code available to the community at \url{URL}\footnote{URL omitted to comply with TACL policy.}.
 We propose a machine-friendly representation of the policy that moderators wish to enforce, by breaking it down into a collection of intents and slots. We collect and annotate a dataset of 3,535 English posts with such slots, and show how architectures for intent classification and slot filling can be used for abuse detection, while providing a rationale for model decisions.\footnote{Accepted at TACL. Our code and data are available at \url{https://github.com/Ago3/PLEAD}.}
\end{abstract}

\section{Introduction}
% 1. Existing models aren’t robust (random AAA scores)
%     1. we should evaluate LFTW model with AAA
% 2. Claim: Models are doing what they are asked, the issue is with the problem formulation
%     1. Humans with the same interpretation of a sentence could still make 2 different decisions (where to draw the line?)
% 3. We re-formulate the problem in a way that humans with same interpretation of the sentence would make the same decision.
%     1. Use the annotation guidelines to draw the line, and provide those guidelines explicitly to the model.
%     2. “Given a post p, Is p abusive?” → “Given a post p and a policy p1, does p violate p1?”

The central goal of online content moderation is to offer users a safer experience by taking actions against abusive behaviours, such as  hate speech. Researchers have been developing supervised classifiers to detect hateful content, starting from a collection of posts known to be abusive and non-abusive.  To successfully accomplish this task, models  are expected to learn  complex concepts from previously flagged examples. For example, hate speech has been defined as ``abusive speech targeting specific group characteristics, such as ethnic origin, religion, gender or sexual orientation'' \cite{2012-warner-detecting}, but there is no clear definition of what constitutes \textit{abusive speech}.

Recent research \cite{2018-dixon-bias} has shown that supervised
models fail to grasp these complexities; instead, they exploit
spurious correlations in the data, they become overly reliant on
\mbox{low-level} lexical features and flag posts because of, for
instance, the presence of group identifiers alone
(e.g.,~\textsl{women} or \textsl{gay}).
%
%Clearly, the same words and phrases can be abusive within one culture or community and harmless in another. For example, reclaimed slurs like \textit{n*gga} and \textit{q*eer} are not abusive when used as such, but are considered hate speech when used by users who do not belong to the corresponding protected communities. To safely evaluate the level of danger of a post, its moderation is usually assigned to human moderators with expertise in the relevant linguistic and societal context.
Efforts to mitigate these problems focus on regularization,
e.g.,~preventing the model from paying attention to group identifiers
during training \cite{2020-kennedy-debias,2020-Zhang-debias}, however,
they do not seem effective at producing better classifiers
\cite{2021-calabrese-aaa}. %, and the use of such techniques seems to suggest that 1) classifiers are not exploiting the input resources properly, and 2) paying attention to group identifiers when deciding about the abusiveness of a post is incorrect.
Social media companies, on the other hand, give moderators detailed
guidelines to help them decide whether a post should be deleted, and
these guidelines also help ensure consistency in their decisions (see
Table~\ref{tab:thought-experiment}). Models are not given access to
these guidelines, and arguably this is the reason for many of their
documented weaknesses.

Let us illustrate this with the following example. Assume we are shown
two posts, the abusive ``\textsl{Immigrants are parasites}'', and the
non-abusive ``\textsl{I love artists}'', and are asked to judge
whether a new post ``\textsl{Artists are parasites}'' is
abusive. While the post is insulting, it does not contain hate speech,
as professions are not usually protected, but we  cannot know that
without access to moderation guidelines. Based on these two posts
alone, we might struggle to decide which label to assign.
%
%Although this training set is small, the scenario is optimal since there is a huge overlap in linguistic content between the labelled examples and the post to classify. Data-centric approaches suggest that you should be able to make a decision, but humans would likely struggle to find consensus on the label to assign.
%
We are then given more examples, specifically the non-abusive
``\textsl{I hate artists}'' and the abusive ``\textsl{I hate
  immigrants}''. In the absence of any other information, we would
probably label the post ``\textsl{Artists are parasites}'' as
non-abusive. %, but to reach such conclusion the main feature they rely on is group identifiers (\textit{immigrants} and \textit{artists}).
The example highlights that 1)~the current problem formulation
(i.e.,~given post~$p$ and a collection of labelled examples~$C$,
decide whether~$p$ is abusive) is not adequate, since even humans
%that agree on the interpretation of $p$
would struggle to agree on the correct classification, and 2)~relying
on group identifiers is a natural consequence of the problem
definition, and often not incorrect.  Note that the difficulty does
not arise due to the lack of data annotated with real moderator
decisions who would be presumably making labeling decisions according
the policy. Rather, models are not able to distinguish between
necessary and sufficient conditions for making a decision based on
examples alone \cite{2022-balkir-necessity}.

\begin{table}[t]
\centering
%\resizebox{0.48\textwidth}{!}{%
\small
\begin{tabular}{@{}p{7.6cm}@{}}
\toprule
\textbf{Post:} Artists are parasites\\
\midrule
\textbf{Policy:} Posts containing dehumanising comparisons targeted
to a group based on their protected characteristics violate the
policy. Protected characteristics include race, ethnicity, national
origin, disability, religious affiliation, caste, sexual
orientation, sex, gender identity, serious disease and immigration
status. \\
\midrule
\textbf{Old Formulation:} Is the post abusive?  \\
\textbf{Our Formulation:} Does the post violate the policy?  \\
\bottomrule
\end{tabular}
\caption{
  While it is hard to judge whether a post is abusive based solely on
  its content, taking the policy into account facilitates decision
  making. The example is based on the Facebook Community
  Standards. 
\label{tab:thought-experiment}}
\end{table}

In this work we depart from the common approach that aims to mitigate undesired model behaviour by adding artificial constraints
% that contradict human experience
(e.g.,~ignoring group identifiers when judging hate speech) and
instead re-define the task through the the concept of
\textit{policy-awareness}: given post~$p$ \emph{and} policy~$P$,
decide whether~$p$ violates~$P$. This entails models are given
policy-related information in order to classify posts like
\textsl{``Artists are parasites''}; e.g.,~they know that posts
containing dehumanising comparisons targeted to a group based on their
protected characteristics violate the policy, and that profession is
not listed among the protected characteristics (see
Table~\ref{tab:thought-experiment}). %In the context of policy-aware abuse detection, humans who agree on the interpretation of $p$ also agree on the classification outcome.
To enable models to exploit the policy, we formalize the task as an
instance of intent classification and slot filling and create a
\emph{machine-friendly representation} of a policy for hate speech by
decomposing it into a collection of intents and corresponding
slots. For instance, the policy in Table~\ref{tab:thought-experiment}
expresses the intent ``Dehumanisation'' and has three slots:
``target'', ``protected characteristic'', and ``dehumanising
comparison''. All slots must be present for a post to violate a
policy. Given this definition, the post in
Table~\ref{tab:thought-experiment} contains a target
(\textsl{``Artists''}) and a dehumanising comparison (\textsl{``are
  parasites''}) but does not violate the policy since it does not have
a value for protected characteristic.

We  create and make publicly available the  \textbf{P}o\textbf{l}icy-aware \textbf{E}xplainable \textbf{A}buse \textbf{D}etection (\textsc{Plead}) dataset which contains (intent and slot) annotations for $3,535$ abusive and non-abusive posts.
To decide whether a post violates the policy and explain the decision, %we use intent classification and slot filling, where slots correspond to properties (e.g., target) and intents to policy rules (e.g., dehumanisation) plus an additional intent representing not-hateful posts. \textbf{W
we design a sequence-to-sequence model that generates a structured representation of the input by first detecting and then filling slots. Intent is assigned deterministically based on the filled slots, leading to the final abusive/non-abusive classification. Experiments show our model is more reliable than classification-only approaches, as it delivers transparent predictions.% and has the additional benefit of producing human readable explanations.

% Our contributions are as follows:
% \begin{itemize}
%     \item We introduce the better defined task of policy-aware abuse detection;
%     \item We create a machine-friendly representation of a policy for hate speech by decomposing it into a conjunction and disjunction of atomic properties;
%     \item We collect an English dataset for explainable policy-aware abuse detection;
%     \item We present a model to address the task in an explainable and robust fashion by reducing the problem of policy-aware abuse detection to an instance of intent classification and slot filling.
% \end{itemize}

%%% Local Variables: 
%%% mode: latex
%%% TeX-master:  "../tacl2021v1-template"
%%% End:

\section{Related Work}
\begin{table*}[t]
\centering
\resizebox{\textwidth}{!}{%
\begin{tabular}{lll}
\toprule
\multicolumn{1}{c}{\bf Intent} & \multicolumn{1}{c}{\bf Policy
  Definition}  & \multicolumn{1}{c}{\bf Examples and Slot Annotations}\\
\midrule
Dehumanisation & \multirow{3}{*}{\shortstack[l]{Posts containing
    \uwave{dehumanising  comparisons} \\targeted to an individual or group
    based on \\their protected characteristics violate the policy.}} &
\multirow{3}{*}{\shortstack[l]{
Immigrants are parasites \\
$[$\texttt{Target}\hspace{.2cm} \textit{Immigrants}$]$,\hspace{.2cm}
$[$\texttt{ProtectedCharacteristic}\hspace{.2cm}
\textit{Immigrants}$]$, \\
$[$\texttt{DehumanisingComparison}\hspace{.2cm} \textit{are parasites}$]$}}\\
\multicolumn{3}{c}{} \\
&& \\
\midrule
Threatening & \multirow{3}{*}{\shortstack[l]{Posts containing
    \uwave{threatening language} \\\uwave{targeted} to an individual or group based on
    \\their \uwave{protected characteristics} violate the policy.}} &
\multirow{3}{*}{\shortstack[l]{"Women should be shot? Agreed!"*
    \\$[$\texttt{Target}\hspace{.2cm} \textit{Women}$]$,\hspace{.2cm}
    $[$ \texttt{ProtectedCharacteristic}\hspace{.2cm}
    \textit{Women}$]$,\\ $[$\texttt{ThreateningSpeech}\hspace{.2cm} \textit{should be shot}$]$}}\\
\multicolumn{3}{c}{} \\
&  & \\
\midrule
Derogation & \multirow{3}{*}{\shortstack[l]{Posts containing
    \uwave{derogatory terms or insults} \\\uwave{targeted} to an individual or group
    based on \\their \uwave{protected characteristics} violate the policy.}} &
\multirow{3}{*}{\shortstack[l]{"Truly, I have been to Poland and there's nothing to see there.
       Polish people are so nasty as well."
       \\$[$\texttt{Target}\hspace{.2cm} \textit{Polish people}$]$,\hspace{.2cm}  $[$\texttt{ProtectedCharacteristic}\hspace{.2cm} \textit{Polish}$]$, \\ $[$\texttt{DerogatoryOpinion}\hspace{.2cm} \textit{are so nasty}$]$}}\\
\multicolumn{3}{c}{} \\
& & \\
\midrule
Animosity & \multirow{3}{*}{\shortstack[l]{Posts containing
    \uwave{\textit{implicit} abusive language} \\\uwave{targeted} to an individual
    or group based on\\ their \uwave{protected characteristics} violate the policy.}} & \multirow{3}{*}{\shortstack[l]{"I don't want to have to be around a breastfeeding woman, if she
       wants to do that she can do it at home"
       \\$[$\texttt{Target}\hspace{.2cm} \textit{a breastfeeding
         woman}$]$,\hspace{.2cm}  $[$\texttt{ProtectedCharacteristic}
       \textit{she}$]$, \\ $[$\texttt{NegativeOpinion}\hspace{.2cm} \textit{can do it at home}$]$}}\\
\multicolumn{3}{c}{} \\
&  & \\
\midrule
Pro Hate Crime & \multirow{2}{*}{\shortstack[l]{Posts that glorify, \uwave{support} or deny \uwave{hateful}\\ actions, events, organisations and \uwave{individuals}.}} & \multirow{2}{*}{\shortstack[l]{ "Hitler was right all along. We are witnessing it at home EVERY
       day." \\$[$\texttt{HateEntity}\hspace{.2cm} \textit{Hitler}$]$,\hspace{.2cm}  $[$\texttt{Support}\hspace{.2cm} \textit{was right all along}$]$}}\\
\multicolumn{3}{c}{}\\
%& & & \\
\bottomrule
\end{tabular}%
}
\caption{
Definition of policy guidelines, intents, and slots associated with
them. Example posts and their annotations. % Note that
%\texttt{ProtectedCharacteristic} models the phrase ``\textsl{based on
 % their protected characteristics}'', and therefore encodes a
%causality effect.
Wording in the guidelines which is mapped onto slots
is \uwave{underlined}.
\label{tab:policy-modelling}}
\end{table*}

We use \textit{abuse} as an umbrella term covering any kind of harmful
content on the Web, as this is accepted practice in the field
\cite{2019-vidgen-challenges,2017-waseem-understandingabuse}. Abuse is
hard to recognise, due to ambiguity in its definition and differences
in annotator sensitivity \cite{2016-ross-measuring}. Recent research
suggests embracing disagreements by developing multi-annotator
architectures that capture differences in annotator perspective
\cite{2022-davani-disagreement,2021-basile-disagreement,2021-uma-disagreement}. While
this approach better models how abuse is perceived, it is not suitable
for content moderation where one has to decide whether to remove a
post and a prescriptive paradigm is preferable
\cite{2022-rottger-prescriptive}.
 %a decision needs to be made about the removal of a post has to be made, as it is the case for content moderation.

\citet{2020-zufall-legal} adopt a more objective approach, as they aim
to detect content that is \textit{illegal} according to EU
legislation. However, as they explain, illegal content constitutes
only a tiny portion of abusive content, and no explicit knowledge
about the legal framework is provided to their model. The problem is
framed as the combination of two binary tasks: whether a post contains
a protected characteristic, and whether it incites violence. The
authors also create a dataset which, however, is not publicly
available.

Most existing work ignores these annotation difficulties and models
abuse detection with transformer-based models
\cite{2021-vidgen-lftw,2020-kennedy-debias,2019-mozafari-bert}. Despite
impressive F1-scores, these models are black-box and not very
informative for moderators. Efforts to shed light on their behaviour,
reveal that they are good at exploiting spurious correlations in the
data but unreliable in more realistic scenarios
\cite{2021-calabrese-aaa,2021-rottger-hatecheck}. Although
explainability is considered a critical capability
\cite{2019-mishra-survey} in the context of abuse detection, to our
knowledge, \citet{2021-sarwar-neighbourhood} represent the only
explainable approach. Their model justifies its predictions by
returning the $k$ nearest neighbours that determined the
classification outcome. However, such ``explanations'' may not be
easily understandable to humans, who are less skilled at detecting
patterns than transformers \cite{2017-vaswani-attention}.

In our work, we formalize the problem of policy-aware abuse detection
as an instance of intent classification and slot filling (ICSF), where
slots are properties like ``target'' and ``protected characteristic''
and intents are policy rules or guidelines (e.g.,
``dehumanisation''). While \citet{2020-Ahmad-privacy} use ICSF to
parse and explain the content of a privacy policy, we are not aware of
any work that infers policy violations in text with
ICSF. State-of-the-art models developed for ICSF are
sequence-to-sequence transformers built on top of pretrained
architectures like BART \cite{2020-Aghajanyan-conversational}, and
also  represent the starting point for our modeling approach.

\section{Problem Formulation}
\label{sec:formulation}

%classification problem

Given a policy for the moderation of abusive content, and a post~$p$,
our task is to decide whether $p$~is abusive. We further note that
policies are often expressed as a set of guidelines $R = \{r_1, r_2, \ldots
r_N\}$ as shown in Table~\ref{tab:policy-modelling} and a post $p$ is abusive when its content
violates any $r_i \in R$. Aside from deciding whether a guideline has been
violated, we also expect our model to return a human-readable explanation which should be specific to $p$ (i.e., an extract from the policy describing the guideline being violated is not an explanation), since customised explanations can help moderators  make more informed decisions, and developers better understand  model behaviour.

%violated by $p$ if any, \texttt{not abusive} otherwise. This is a classification problem with $N + 1$ classes.
%explainable problem

%We further define the problem of \textbf{Policy-Aware Explainable Abuse Detection}, where on top of the classification label the model is expected to return 

%\section{Explainable Abuse Detection as Intent Classification and Slot Filling}
%\label{sec:policy}
% 4. First explain what our approach is (i.e., why we need to decompose the policy, we could already answer the question above keeping the whole policy text as input)
%     1. Write about explainability and everything
% 5. Policy choice and modelling:
%     1. AT policy: non-commercial, but similar to Hate Speech sections in Social Media policies
%     2. Decompose policy into atomic rules
%     3. Model rules as a conjunction of conditions

\paragraph{Intent Classification and Slot Filling}
The generation of post specific explanations requires detection
systems to be able to reason over the content of the policy. To
facilitate this process, we draw inspiration from previous work
\cite{2018-gupta-hierarchical} on intent classification and slot
filling (ICSF), a task where systems have to classify the intent of a
query (e.g.,~\texttt{IN:CREATE\_CALL} for the query \textit{``Call
  John''}) and fill the slot associated with it (e.g.,
\textit{``Call''} is the filler for the slot \texttt{SL:METHOD} and
\textit{``John''} for \texttt{SL:CONTACT}). For our task, we decompose
policies into a collection of intents corresponding to the guidelines
mentioned above, and each intent is characterized by a set of
properties, i.e.,~{slots} (see Table~\ref{tab:policy-modelling}).

\begin{figure}[t]
    \centering
    \includegraphics[width=0.48\textwidth]{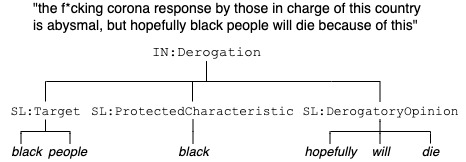}
    \caption{Decoupled representation for a post.}
    \label{fig:representation}
\end{figure}

The canonical output of  ICSF systems is a tree
structure. Multiple representations have been defined, each with a
different trade-off between expressivity and ease of parsing. For our
use case, we adopt the decoupled representation proposed in
\citet{2020-Aghajanyan-conversational}: non-terminal nodes are either
slots or intents, the root node is an intent, and terminal nodes are
words attested in the post (see Figure~\ref{fig:representation}). In this
representation, it is not necessary for all input words to appear in
the tree (i.e., in-order traversal of the tree cannot reconstruct the
original utterance). Although this ultimately renders the 
parsing task harder, it is crucial for our domain where words can be
associated with multiple slots or no slots, and reasoning over long-term
dependencies is necessary to recognise, e.g., a derogatory opinion
(see Figure~\ref{fig:representation}).

Importantly, we first identify the slots occurring in a post and then
deterministically infer the author's intent, as this renders the
output {tree} an \textit{explanation} of the final classification
outcome rather than a {post-hoc} \textit{justification}
\cite{2017-biran-explanation}. Likewise, since we view the predicted
slots as an explanation for intent, we cannot jointly perform intent
classification and slot filling, to avoid producing inconsistent
explanations \cite{2020-camburu-expl,2022-ye-unreliability}.

\paragraph{Hate Speech Taxonomy}
As a case-study, we model the
codebook\footnote{\url{https://github.com/bvidgen/Dynamically-Generated-Hate-Speech-Dataset}}
for hate speech annotations designed by the Alan Turing Institute
\cite{2021-vidgen-lftw}. This policy is very similar to the guidelines
that social media platforms provide to moderators and
users.\footnote{e.g.,
  \url{https://transparency.fb.com/en-gb/policies/community-standards/hate-speech}}

We obtained an intent from each section of the policy, and associated
it with a set of slots (see Table~\ref{tab:policy-modelling}). We
followed the policy guidelines closely and slots were mostly extracted
verbatim from them (see underlined policy terms in
Table~\ref{tab:policy-modelling} which give rise to slots). We
refrained from renaming or grouping slots to create more abstract
labels (e.g., using \texttt{SL:AbusiveSpeech} to replace
\texttt{SL:Dehu\-ma\-ni\-sing\-Com\-pa\-ri\-son},
\texttt{SL:Threa\-te\-ning\-Speech},
\texttt{SL:De\-ro\-ga\-to\-ry\-O\-pi\-nion}, and
\texttt{SL:Ne\-ga\-ti\-ve\-O\-pi\-nion}).  Note that commonsense
knowledge is required to decide whether a span is the right filler for
a slot. For instance, [\texttt{SL:ThreateningSpeech}\hspace*{.2cm}\textit{dog}] would be odd, while
[\texttt{SL:ThreateningSpeech}\hspace*{.2cm}\textit{should be shot}]
wouldn't.

In addition to slots
corresponding to different types of hate speech, most intents have
a \texttt{Target} who is being abused because of a
\texttt{ProtectedCharacteristic}.  In contrast to previous work
\cite{2020-sap-sbic,2019-ousidhoum-mlma}, we distinguish targets from
protected groups, as this allows annotators to better infer the
target's characteristics from context.  A post is deemed abusive
(i.e., violates the policy) if and only if all slots for at least one
of the (hateful) intents are filled. We also introduce a new intent (i.e., \texttt{IN:NotHateful}) to accommodate all  posts that do not violate the policy.

% \begin{equation}
%     r \iff \forall P \in \delta(r_i) \enspace \exists \, s \in S(p): \, P(s)
% \end{equation}
% where $\delta(r)$ is the set of relations in the modelling of $r$, and $S(p)$ is the set of non-empty subsequences of $p$ (see Table~\ref{tab:policy-modelling}).

% If $r$ holds (i.e. if a non-empty subsequence was found for each property such that all properties hold true), then the post $p$ violates the rule $r$.

Besides being more machine-friendly, our formulation is advantageous
in reducing the amount of abusive instances required for training,
since a model can learn to predict slots even from non-abusive
instances (e.g., slots \texttt{SL:Target} and
\texttt{SL:DehumanisingComparison} are also present in the non-abusive
\textsl{``Artists are parasites''}). This is particularly important in
this domain, since in absolute terms, abusive posts are (luckily)
relatively infrequent compared to non-abusive ones
\cite{2018-founta-abusive}, and most harmful content is detected by
moderators and subsequently deleted.

\paragraph{Counter Speech} In a few cases, posts might quote hate
speech, but the authors clearly distance themselves from the harmful
message. To enable models to correctly recognise counter speech ---
speech that directly counters hate, for example by presenting facts or
reacting with humour \cite{2019-mathew-thou} --- we introduce a new
slot encoding the author's stance
(i.e.,~\texttt{SL:NegativeStance}). For instance, the post
\textsl{``It's nonsense to say that Polish people are nasty''}
expresses a derogatory opinion which is based on a protected
characteristic of a target (i.e., ``\textsl{Polish people}''). Even
though all slots for the \texttt{Derogation} intent are filled, the
post is not abusive as the author is reacting to the hateful message.
A post is hateful if and only if there are fillers for all associated
slots but not for \texttt{SL:NegativeStance}.
% $$r_i \iff r_i \land \neg \, NegativeStance(x)$$
% \begin{align}
% \begin{split}
%     \label{eq:classification}
%     r \iff & \forall P \in \delta(r) \enspace \exists \, s \in S(p): \, P(s) \enspace \land  \\ & \nexists \, s  \in S(p): \, NegativeStance(s)
% \end{split}
% \end{align}

 %This enables models to recognise \textit{``It's nonsense''} as the
%author's negative stance, and therefore to correctly classify the post
%as not hateful.

%%% Local Variables: 
%%% mode: latex
%%% TeX-master:  "../tacl2021v1-template"
%%% End:

\section{The \textsc{Plead} Dataset}
\label{sec:data}
\begin{table*}[t]
\centering
\resizebox{\textwidth}{!}{%
\begin{tabular}{l|rcc|rcc|rcc|rcc|rcc}
\toprule
& \multicolumn{3}{c|}{\textbf{Dehumanisation}} & \multicolumn{3}{c|}{\textbf{Threatening}} & \multicolumn{3}{c|}{\textbf{Derogation}} & \multicolumn{3}{c|}{\textbf{Pro Hate Crime}} & \multicolumn{3}{c}{\textbf{Not Hateful}}\\
& \textbf{N\textdegree{}} & \textbf{LCS (\%)} & \textbf{A (\%)} & \textbf{N\textdegree{}} & \textbf{LCS (\%)} & \textbf{A (\%)} & \textbf{N\textdegree{}} & \textbf{LCS (\%)} & \textbf{A (\%)} & \textbf{N\textdegree{}} & \textbf{LCS (\%)} & \textbf{A (\%)} & \textbf{N\textdegree{}} & \textbf{LCS (\%)} & \textbf{A (\%)}\\
\midrule
{Target} & 972 & 63.38 & 97.32 & 610 & 71.46 & 98.60 & 1102 & 64.91 & 97.91 & 0 & --- & --- & 836 & 56.75 & 95.16 \\
{ProtectedCharacteristic} & 1006 & 75.50 & 95.02 & 639 & 82.44 & 97.08 & 1156 & 78.63 & 95.16 & 0 & --- & --- & 139 & 68.45 & 93.60 \\
{DehumanisingComparison} & 883 & 50.99 & 96.32 & 0 & --- & --- & 0 & --- & --- & 0 & --- & --- & 36 & 60.42 & 97.22 \\
{ThreateningSpeech} & 0 & --- & --- & 585 & 56.89 & 97.55 & 0 & --- & --- & 0 & --- & --- & 48 & 51.45 & 96.63 \\
{DerogatoryOpinion} & 0 & --- & --- & 0 & --- & --- & 994 & 45.56 & 93.50 & 0 & --- & --- & 378 & 48.97 & 92.98 \\
{HateEntity} & 0 & --- & --- & 0 & --- & --- & 0 & --- & --- & 173 & 69.87 & 94.64 & 1 & 100.00 & 33.33 \\
{Support} & 0 & --- & --- & 0 & --- & --- & 0 & --- & --- & 173 & 44.44 & 90.71 & 0 & --- & --- \\
{NegativeStance} & 0 & --- & --- & 0 & --- & --- & 0 & --- & --- & 0 & --- & --- & 40 & 53.87 & 94.02 \\
\midrule
Number of instances & \multicolumn{3}{c|}{883} & \multicolumn{3}{c|}{585} & \multicolumn{3}{c|}{994} & \multicolumn{3}{c|}{173} & \multicolumn{3}{c}{900} \\
%\textbf{N\textdegree{} of one-target instances} & \\
\bottomrule
\end{tabular}
}
\caption{Number of occurrences per slot for each intent;
  inter-annotator agreement measured by Longest Common Subsequence
  score (LCS), and percentage of annotations approved by expert (A).
\label{tab:dataset}}
\end{table*}%
\paragraph{Post selection} To validate our
problem formulation and for model training we created a dataset
consisting of posts with slot annotations (e.g.,~\texttt{Target},
\texttt{ThreateningSpeech}). We built our annotation effort on an
existing dataset associated with the policy guidelines introduced in
Section~\ref{sec:formulation} and extended it with additional
span-level labels. This dataset \cite{2021-vidgen-lftw} was created by
providing annotators with a classification model trained on 11 other
datasets, and asking them to write hateful and non-hateful sentences
such that they fooled the model in predicting the opposite class
(hateful for non-hateful and vice versa). The process was iterative,
we used sentences from the second round onwards, which were annotated
with policy violations.
%The process was repeated iteratively 4 times, each time training the model with the additional data produced in the previous rounds.
%While in the first round instances were annotated only with a binary hateful/not-hateful label, hateful sentences produced in the following rounds are also associated with the violated rule.

 The dataset is not balanced, i.e.,~some policies are violated more
 frequently than others.  To mitigate this and reduce annotation
 costs, we selected all posts from the less popular policies and a
 random sample of posts from the most popular ones.  We further merged
 posts annotated with derogation and animosity classes as they are
 similar, the main difference being the extent to which the negative
 opinion is implied.  The number of selected posts per intent is shown
 in Table~\ref{tab:dataset}. We note that this is a collection of hard
 examples, as they were written so as to fool a state-of-the-art
 model.  Most non-abusive posts in the dataset have annotations for
 all slots save one, or they contain counter speech and are easily
 confusable with hate speech. 
 
% rarest class including only ${\sim}{200}$ posts. Furthermore, we note

% We need to collect training data where all the properties are annotated
% We start from LFTW because:
% - it contains rule-level annotations compatible to the policy above (for rounds other than 1)
% - it's a dataset of hard examples: written to trick the models (many non abusive posts only have one element missing or are counter speech)
% - compatibility with CAD, so possibility to extend to another dataset without need for new annotations
% Table showing how many instances we selected (for smaller categories, all except the ones we used to train the annotators)
%
% \begin{figure}
%     \centering
%     \includegraphics[width=0.5\textwidth]{figures/example_annotation.png}
%     \caption{Annotation task for Dehumanisation posts.}
%     \label{fig:annotation}
% \end{figure}
%
\paragraph{Annotation Task} We performed two annotation tasks, one for
hateful posts and one for non-hateful ones.  For hateful posts,
annotators were presented with the post, information about the
target(s), its characteristics, and the slots. They were then asked to
specify the spans of text corresponding to each slot. The dataset
already contains annotations about which policy is being violated.
For instance, for posts labelled as \texttt{Pro Hate Crime},
annotators look for spans corresponding to \texttt{HateEntity} and
\texttt{Support}. Information about the target and its characteristics
is also present in metadata distributed with the dataset, and we used
it to steer annotators towards a correct reading of the posts. In
general the original posts, metadata, and labels are of high-quality;
\citet{2021-vidgen-lftw} report extremely high agreement for instances
created during round 2, moderate for the following rounds and
disagreements were resolved by an expert annotator.
  
Each post can contain multiple targets, and each target can be
associated with multiple protected characteristics (e.g.,
\textsl{black woman} indicates both the race and gender of a
target). Our annotation scheme assumes that only one opinion is
annotated for each post. For instance, the post ``\textsl{I love black
  people but hate women}'' contains both a non-hateful and hateful
opinion, but we only elicit annotations for the hateful one. Likewise,
when a post contains more than one hateful opinion\footnote{Manual
inspection of a sample of hateful instances revealed the percentage of
instances with multiple hateful opinions to be~$\sim{3}\%$.},
annotators select the one that better fits the associated policy and
target description.  Equally, for non-hateful posts, we asked
annotators to focus on a single opinion, with a preference for
opinions that resemble hateful messages (e.g.,~the second opinion in
``\textsl{I love cats, but I wish all wasps dead}''). Annotators could
specify as many spans (and associated slots) as they thought
appropriate, including none. If enough elements were selected for a
post to violate a rule (e.g., both \texttt{HateEntity} and
\texttt{Support} were specified), annotators were asked whether the
post contained counter speech (and if so, to specify a span of text
for \texttt{NegativeStance}) or derogatory terms used as reclaimed
identity terms (e.g., the n-word used by a member of the Black
community).

% Describe the annotation task:
% - We want to collect data to learn the properties, and we add some fine-grained questions because you never know
% - we ask for attributes for protectedcharacteristics and comparisons, and for stance
% - for hateful posts, they are just asked to fill-in the slots (also number of targets and attribute is provided), and specify if the author is supporting someone else's hateful message
% ---- this implies an assumption -> maybe there were multiple opinions, but they only had to focus of the specified one
% - for not-hateful posts, we want annotators to focus on only one opinion (to be coherent with the other task), so we ask them to focus on the part of the post that reminds them about hateful posts the most (e.g., parts where there are 2 slots out of a 3-slots rule. They all worked on at least one hateful task).
% ---- They are free to choose which slots occurr (even none), and then are asked to specify if it's counter speech (indicated with a negative stance) or reclaimed.
%Reclaimed instances are too infrequent and not covered from the modelling above. We decide to discard these instances, as there is not enough data to learn an ad-hoc defined relation.

\paragraph{Annotator Selection} We recruited annotators resident in
English-speaking countries through the Amazon Mechanical Turk
crowdsourcing platform. To ensure high-quality annotations we designed
a quiz for each policy rule and assessed the fairness of the quiz
through a two-phase pilot study: in the first phase annotators were
shown the instructions and asked to annotate eight sentences. These
annotations were then used as possible correct answers for the quiz or
to clarify the instructions. During the second phase, \textit{new}
annotators were shown the updated instructions and asked to pass a
quiz consisting of three questions.

The pilot showed that most crowdworkers who understood the task were able to
pass the quiz, but no one was able to pass the quiz without
understanding the task. %, and so we considered that the quiz was
%successful.
Only successful annotators were granted a \emph{guideline-specific}
qualification that allowed them to annotate real instances. To enforce
consistency, annotators were prompted to pass a quiz after
every 30-post batch, and each batch contained posts associated
with one rule. To ensure the data was annotated correctly, we included
two control questions in each batch. These were not simple attention
checks, but regular posts for which the correct answers were known
(from the pilot study).
%To minimise confusion, annotators were prompted to work on the same rule until no more sentences from that category were available. After that, they were able to apply for a new qualification. 
For the annotation of non-hateful posts, we only admitted annotators
who had submitted at least 300 annotations for hateful posts, and
used the first batch of annotations as a further qualification test.
%During both annotation tasks, submissions were constantly monitored and annotators continuously provided feedback. When annotation quality decreased for an annotator, they were invited to retake the qualification round.
Overall, 75\% of annotations were produced by women, 91\% by people
who identify as straight, and 75\% by people with ethnicity other than
white. We will release a full breakdown of demographic information
with our dataset.
%Platform used and how we select workers (both geographical criteria and obscene content and quiz for qualification). One qualification per rule. Quiz designed with a 2-round trial study. If you pass the 3-questions quiz, you get a qualification but
%every time you still need to pass a 1-question quiz (to enforce consistency). This means that each batch only contains posts from one rule, and workers are prompted to keep working on the same category till exhaustion of sentences. Each HIT contains 2 control questions (not attention check, more a normal annotation for which we know the answer) to check the worker is really doing the task.
% Around 20% of each hit is monitored constantly, and annotators are continuously provided feedback. For workers that produce lower-quality annotations, qualification was removed. For not-hateful instances, there is the additional requirement of having submitted at least N HITs on the hateful task, and there is no quiz but all instances are checked manually in the first HIT or couple of HITs, depending on the cases. Overall we had N workers working on the hateful task, and M on the not-hateful tasks. Each worker worked on avg on R tasks. Demographic info about the workers can be found in the appendix.

\paragraph{Annotation Quality} We collected three
annotations per post. %\taclrevisions{Common practice in the creation of datasets for intent classification and slot filling is to measure annotation quality by having a further annotator reviewing the annotations \cite{}.}
%All annotations are released, together with anonymised
%demographic information of the corresponding annotators.
We measured inter-annotator agreement, as the length of the longest
common subsequence (LCS) normalized by the length of the longest
span. LCS better reflects our annotation task compared to more
standard measures like Cohen's Kappa \cite{McHugh:2012} which could be
calculated on the token level. However, our annotators do not consider
words indvidually, they primarily make decisions for sequences of
tokens. Aside from presuming a different annotation task, token-based
Kappa leads to extremely unbalanced data with the un-annotated tokens
vastly outnumbering the labels of interest,  artificially
inflating agreement.

As shown in Table~\ref{tab:dataset}, average LCS scores are high for
relations like \texttt{Target} and \texttt{ProtectedCharacteristic},
but drop to ${\sim}{45\%}$ for relations like
\texttt{DerogatoryOpinion}. After a quick inspection, we discovered
that poor scores were the result of non overlapping but nevertheless
valid annotations.  For instance, in ``\textsl{Homosexuality means the
  end of the family. The next step is legalizing pedophilia.}'', both
```\textsl{means the end of the family}'' and ``\textsl{legalizing
  pedophilia}'' were annotated as derogatory opinions, but since there
is no overlap between the spans, agreement is 0. To better assess the
quality of the annotations, an expert annotator manually reviewed
$50\%$ in each category. %During
%validation, each span of text was accepted (as is, or slightly
%modified to enforce consistency) or discarded.
As shown in
Table~\ref{tab:dataset}, the percentage of valid annotations was
remarkably high, which leads as to conclude that they are reliable. 
%Annotation agreement. Agreement is low, because of multiple correct answers. We decide to validate half instances for each category.
%Final product: 3 annotations per instance, with demographic info of each annotation (but not associated to a real ID!). Plus examples of what annotation guidelines and quiz questions looked like.

\paragraph{A Note on Ethics} To protect annotators from exposure to
hateful content, we tagged our project with the ``offensive content''
label on Amazon Mechanical Turk, included a warning in the task title, and
asked for consent twice (first at the end of the information sheet,
and then with a one-sentence checkbox). Annotators were presented with
small batches of 30 sentences, and invited to take a break at the end
of each session. They were also offered the option to quit anytime
during the session, or to abandon the study at any point. A reminder
to seek for help in case they experienced distress was provided at the
beginning of each session. The study was approved by the relevant
ethics committee (details removed for anonymous peer review). 

%%% Local Variables: 
%%% mode: latex
%%% TeX-master:  "../tacl2021v1-template"
%%% End:

\section{Abuse Detection Model}
\begin{figure*}[t]
    \centering
    %\begin{subfigure}[]{0.95\textwidth}
\begin{tabular}{l@{~~}c}
  \raisebox{3cm}[0pt]{(a)} &  \includegraphics[width=0.95\textwidth]{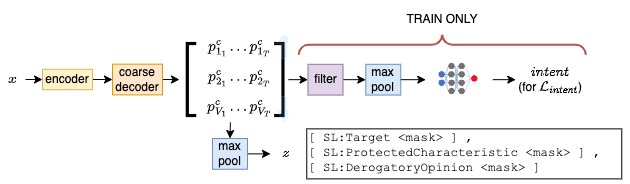}\\
    %\caption{\label{fig:model-a}}
    %\end{subfigure}
    %\begin{subfigure}[]{0.95\textwidth}
\raisebox{1.6cm}[0pt]{(b)}   & \includegraphics[width=0.95\textwidth]{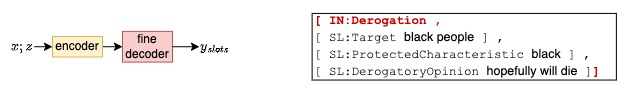}\\
    %\caption{\label{fig:model-b}}
    %\end{subfigure}
    \end{tabular}
    \caption{We first generate the meaning sketch based on the input
      post (a), and then refine it by filling the slots (b). The
      intent (in red) is inferred deterministically based on predicted
      slots~$y_{slots}$. The model is trained with an intent-aware
      loss (a). \label{fig:model}}
\end{figure*}
ICSF is traditionally modelled as a sequence-to-sequence problem where
the input utterance represents the \textit{source} sequence, and
the \textit{target} sequence is a linearised version of the
corresponding tree. For instance, the linearised version of the tree
in Figure~\ref{fig:representation} would be: [\texttt{IN:Derogation},
[\texttt{SL:Target}, \textit{black, people}],
[\texttt{SL:ProtectedCharacteristic}, \textit{black}], \ldots ]. Due
to the nature of our domain, where posts can contain multiple
sentences all of which might have to be considered to discover policy
violations (e.g., because of coreference), we adopt
the \textit{coversational} approach to ICSF introduced
in \citet{2020-Aghajanyan-conversational}. In this setting, all
sentences are parsed in a single session (rather than
utterance-by-utterance) which is pertinent to our task, as we infer 
intent \textit{after} filling the slots, and would otherwise have no
information on which slots to carry over (e.g., detecting a target in
the first utterance does not constrain the set of slots that could
occur in the following ones).

% \begin{figure*}
%     \centering
%     \includegraphics[width=\textwidth]{figures/model.jpg}
%     \caption{Our model first generates the meaning sketch from the input post, and then refines it by filling the slots.}
%     \label{fig:model}
% \end{figure*}
%
% \begin{figure*}
%     \centering
%     \includegraphics[width=\textwidth]{figures/model_one_per_line.jpg}
%     \caption{We first generate the meaning sketch from the input post (a), and then refine it by filling the slots (b). Intent is selected deterministically based on the slots (c). The model is trained with an intent-aware loss (d).}
%     \label{fig:model}
% \end{figure*}
%

Our sequence-to-sequence model is built on top of BART
\cite{2020-lewis-bart}.
However, %we modify the architecture to better suit our use case. In fact,
in canonical ICSF, BART generates the intent first, and then uses it
to look for the slots associated with it. In our case, intent is
inferred post-hoc, based on the identified slots, not vice versa. Our
model adopts a two-step approach where BART first generates a coarse
representation of the input, namely a meaning sketch with
coarse-grained slots, and then refines it
\cite{2018-dong-meaningsketches}. The meaning sketch is a tree where
non-terminal nodes are slots, and leaves are \texttt{<mask>} tokens.
The sketch for the example in Figure~\ref{fig:representation} and its
refined
version %would be: [\texttt{SL:Target}, \textit{<mask>}], [\texttt{SL:ProtectedCharacteristic}, \textit{<mask>}], [\texttt{SL:DerogatoryOpinion}, \textit{<mask>}].
are shown in Figure~\ref{fig:model}.  Specifically, we first encode
 source tokens $w_i$ (Figure~\ref{fig:model}a): 
$$e_1, \ldots, e_{|x|} = \operatorname{Encoder}(w_1, \ldots, w_{|p|})$$
where $|x|$ is the number of tokens in  post~$x$, and then use the
hidden states to generate the meaning sketch by computing probability
distribution~$\mathtt{p}^c$ over the vocabulary for each time step $t$
as:
\[
d_t^c = \operatorname{Decoder}_{c}(e_1, \ldots, e_{|x|};d_{t-1}^c;s_{t-1}^c)
\]

\[\mathtt{p}_t^{c} = \operatorname{softmax}(W d_t^c + b)\]
where $s_{t-i}^c$ is the incremental state of the decoder. We then decode the meaning sketch $z_1, \ldots, z_T$:
$$z_t = \operatorname{argmax}(\mathtt{p}_t^c)$$
And refine it (see Figure~\ref{fig:model}b) by first re-encoding the
source tokens jointly with the meaning sketch (which is gold at
training time, predicted otherwise): 
$$v_1, ..., v_{|x| + T} = \operatorname{Encoder}(w_1, ..., w_{|x|}; z_1, ..., z_T)$$
A  second decoder  generates then a new probability distribution
over the vocabulary: 
$$d_t^f = \operatorname{Decoder}_{f}(v_1, \ldots, v_{|x|+T};d_{t-1}^{f};s_{t-1}^f)$$
$$\mathtt{p}_t^{f} = \operatorname{softmax}(W d_t^f + b)$$
At inference time, we use beam search to generate the final representation starting from $\mathtt{p}^{f}$.

The training objective is to jointly learn to generate the correct
sketch~$z$ for post~$x$, and the correct tree~$t$ from~$x$
\emph{and}~$z$. We define our loss function for  tuple~\mbox{$(x, z, t)$} as:
$$L_{c,i} = - \sum_{v \in V} \mathbb{1}_{[z_i = v]} \enspace \log(\mathtt{p}^c_{i, v})$$
$$L_{f,i} = - \sum_{v \in V} \mathbb{1}_{[t_i = v]} \enspace \log(\mathtt{p}^f_{i, v})$$
$$\mathfrak{L} = \operatorname{mean}_{i}(L_{c,i}) + \operatorname{mean}_{i}(L_{f,i})$$
where~$V$ is the vocabulary and~$i$ is an index over the sequence
length.

%\paragraph{Intent-aware loss}

Although this loss penalises the model for hallucinating or missing
slots, it does not discriminate between errors that cause the
prediction of a wrong intent, and those that are less relevant
(e.g.,~hallucinating a threat when no target has been detected). In
fact, intent is not part of our sequence-to-sequence task since it is
only predicted post-hoc.  To help the model learn how combinations of
slots relate to intents, we include intent classification as an
additional training task.

We essentially predict intent starting from the probability of each
slot to appear in the sketch (Figure~\ref{fig:model}a). In other
words, we restrict $\mathtt{p}_t^c$ to slot tokens (e.g.,
\texttt{SL:Target}) and normalise it, to obtain a new probability
distribution~$\mathtt{q}_t$ over the set of slots. We aggregate these
probabilities by taking the maximum value over the sequence length,
thus obtaining a single score for each slot. Since each intent can be
modelled as a disjunction of slot combinations (e.g., the
\texttt{NotHateful} intent could result from a tree containing only a
target, or only a target AND a protected characteristic),
we pass the slot scores through two linear layers with activation
functions:
$$s_{slots} = \operatorname{ReLU}(W_{s2s}\, \mathtt{q} + b_{s2s})$$
$$s_{intent} = \operatorname{softmax}(W_{s2i}\, \mathtt{q} + b_{s2i})$$
thus obtaining a probability distribution $s_{intent}$ over
intents~$I$. $W_{s2s} \in \mathbb{R}^{|S|\times|S|}$ models
slot-to-slot interactions, while
$W_{s2i} \in \mathbb{R}^{|S|\times|I|}$ models interactions between
combinations of slots and intents. We then modify our loss to
include the new classification loss for an input post with intent~$c$:
$$L_{intent} = - \sum_{i \in I} \mathbb{1}_{[c = i]} \enspace \log(s_{intent,i})$$
$$\mathfrak{L} = \operatorname{mean}_{i}(L_{c,i}) + \operatorname{mean}_{i}(L_{f,i}) + L_{intent}$$

The new loss aims to assign higher penalty to meaning sketches that
lead to intent misclassification. The two linear layers are trained on
gold intents and sketches. The layers are then added to the BART-base
architecture while kept frozen, so that the model cannot modify its
weights to ``cover up'' wrong sketches by still mapping them to the
right intents. Note that this additional classification task is only
meant to improve the quality of the generated sketches: intent 
is added post-hoc in the output tree depending on the slots that have
been detected (Figure~\ref{fig:model}a).

%%% Local Variables: 
%%% mode: latex
%%% TeX-master:  "../tacl2021v1-template"
%%% End:

\section{Experimental Results}
%\paragraph{Data}
We performed experiments on the \textsc{Plead} dataset
(Section~\ref{sec:data}). Rather than learning complex structures with
nested slots, we post-process an instance with $T$ targets into
$T$~equivalent instances, one per target. Furthermore, we discarded
instances with reclaimed identity terms as these are not taken into
account by our current modelling of the policy, and are too infrequent
(\mbox{$< 0.01\%$}). We split the dataset into training, validation
and test set (80\%/10\%/10\%), keeping the same intent distribution
over the splits.

\subsection{Why Explainability?}
\begin{table}[t]
\centering
\resizebox{0.4\textwidth}{!}{%
\begin{tabular}{c|rrrrr}
\toprule
\textbf{Seed} & \textbf{F1} & \textbf{AAA} & \textbf{GI$_N$} & \textbf{GI$_P$} & \textbf{IND} \\
\midrule
1 & 79.04 & 52.27 & 58.57 & 49.05 & 72.31 \\
2 & 79.04 & 54.10 & 61.43 & 54.76 & 50.77 \\
3 & 80.45 & 56.70 & 52.86 & 75.71 & 56.92 \\
4 & 80.74 & 47.18 & 34.29 & 62.86 & 56.92 \\
5 & 80.74 & 35.69 & 21.43 & 23.33 & 52.31 \\
\midrule
Std & 0.89 & 8.31 & 17.20 & 19.44 & 8.54 \\
\bottomrule
\end{tabular}}
\caption{
Performance of RoBERTa on PLEAD (measured by F1 and AAA) and HateCheck
functionality tests for neutral  (GI$_N$) and positive (GI$_P$) group identifiers and
attacks on individuals (IND).
\label{tab:roberta-experiment}}
\end{table}%

Our first experiment  provides empirical support for our
hypothesis that classifiers trained on collections of abusive and
non-abusive posts do not necessarily learn representations directly
related to abusive speech. We would further argue that if a model
performs well on the test set, it has not necessarily learned to
detect abuse.  For this experiment, we trained
RoBERTa \cite{2021-vidgen-lftw} with five different random seeds, and
obtained an F1-score of ${\sim}{80}\%$ in the binary classification
setting with a low standard deviation (see
Table~\ref{tab:roberta-experiment}). We further examined the output of
these five RoBERTa models using AAA \cite{2021-calabrese-aaa} and
HateCheck \cite{2021-rottger-hatecheck}.  AAA stands for Adversarial
Attacks against Abuse and is a metric that better captures a model's
performance on hard-to-classify posts, by penalising systems
which are biased on low-level lexical features. It does so by
adversarially modifying the test data (based on patterns found in the training data) to generate
plausible test samples. HateChek is a suite of functional
tests for hate speech detection models.

Firstly, we observe high standard deviations across AAA-scores. Models
obtained with seeds 4 and 5 have identical F1-scores, but a gap of 12
points on AAA, suggesting that they may be modelling different
phenomena. HateCheck tests on group identifiers confirm this
hypothesis, as the model trained with random seed~5 misclassifies most
neutral (GI$_N$) or positive (GI$_P$) sentences containing group
identifiers as hateful, while the model trained with seed~4 can
distinguish between different contexts and recognises most positive
sentences as not hateful. Likewise, the models obtained with seeds 1
and 2 have identical F1-scores, and also similar AAA-scores, but a
20~point gap on the test containing attacks on individuals (IND). This
suggests that classifiers tend to model different phenomena (like the
presence of group identifiers or violent speech) rather than policy
violations and that similarities in terms of F1-score disguise
important differences amongst models.

%if we don't constrain classifiers to model \textit{our}
%phenomenon (i.e., policy violations), then they can take on
%on \textit{any} phenomenon, . Behind the very can hide models
%answering different questions, so if we want to have any control on
%what we release, and we must, we need explainability.

\subsection{Model Evaluation}

Since the output of our model is a parse tree, we represent it as set
of productions and evaluate using F1 \cite{2015-quirk-productionf1}
on: (a)~the entire tree (PF1), (b)~the top level (i.e., productions rooted in
intent, PF1$_I$), and (c)~the lower level (i.e., productions rooted in
correctly detected slots, PF1$_L$).  We also report exact match
accuracy for the full tree~(EMA$_T$). 

We compare our model (BART+MS+I) to ablated versions of itself,
including a BART model without meaning sketches or an intent-aware
loss, and a variant with meaning sketches but no intent-aware loss
(BART+MS).  We also compare against two baselines which encode the
input post with an LSTM or BERT, respectively, and then use a
feed-forward neural network to predict which slot labels should be
attached to each token \cite{2021-weld-icsfsurvey}. The LSTM was initialised with Glove embeddings \cite{2014-pennington-glove}. For BERT, we
concatenate the hidden representation of each token to the embedding
of the \texttt{CLS} token, and compute the slots associated to a word
as the union of the slots predicted for the corresponding subwords.
We enhance these baselines by modeling  slot prediction as a
multi-label classification task (i.e., one-vs-one) in line with
\citet{2020-pawara-onevsone}.  For each pair of slots $<s_1, s_2>$, we
introduce an output node and use  gold label $1$ ($-1$) if $s_1$
($s_2$) is the right tag for the token, and~$0$ otherwise.  

As an upper bound, we report F1 score by comparing the
annotations of one crowdworker against the others.  Recall that
annotation of hateful posts was simplified by asking participants to
look for specific slots; as a result, some scores are only available
for non-hateful instances where annotators could select from all the
slots.

\begin{table*}[t]
\centering
\resizebox{\textwidth}{!}{%
\begin{tabular}{l||rrr|rrr|rrr|rrr||r||rrr}
\toprule
\multirow{3}{*}{\textbf{Model}} & \multicolumn{12}{c||}{\textbf{Tree}} & \textbf{Intent} & \multicolumn{3}{c}{\textbf{Hateful?}}\\
 & \multicolumn{3}{c}{\textbf{PF1}} & \multicolumn{3}{c}{\textbf{PF1$_I$}} & \multicolumn{3}{c}{\textbf{PF1$_L$}} & \multicolumn{3}{c||}{\textbf{EMA$_T$}} & \multicolumn{1}{c||}{\multirow{2}{*}{\textbf{F1}}} & \multicolumn{1}{c}{\multirow{2}{*}{\textbf{F1}}} & \multicolumn{1}{c}{\multirow{2}{*}{\textbf{AAA}}} & \multicolumn{1}{c}{\multirow{2}{*}{\textbf{Mean}}} \\
& \textbf{H} & \textbf{NH} & \multicolumn{1}{r}{\textbf{All}} & \textbf{H} & \textbf{NH} & \multicolumn{1}{r}{\textbf{All}} & \textbf{H} & \textbf{NH} & \multicolumn{1}{r}{\textbf{All}} & \textbf{H} & \textbf{NH} & \textbf{All} & & & & \\
\midrule
Humans & --- & 61.52 & --- & --- & 81.23 & --- & 59.90 & 55.80 & 59.29 & --- & 21.59 & --- & --- & --- & --- & --- \\
\midrule
% RoBERTa & - & - & - & - & - & - & - & - & - & - & - & - & \textbf{69.18} & 80.00 & 49.19 & \textbf{64.60} \\
LSTM & 40.66 & 20.08 & 36.30 & 54.20 & 18.42 & 45.97 & 33.91 & 32.03 & 34.63 & 0.00 & 0.90 & 0.11 & 48.10 & 57.73 & 40.55 & 49.14 \\
{BERT} & 40.45 & 16.58 & 35.26 & 59.54 & 16.92 & 50.31 & 25.39 & 23.43 & 25.02 & 0.00 & 0.00 & 0.00 & 52.35 & 76.77 & 31.19 & 53.98 \\
\midrule
BART & 37.57 & 57.26 & 41.45 & 37.73 & 74.13 & 45.12 & 47.82 & 54.89 & 49.15 & 15.51 & 2.73 & 5.95 & 51.78 & 62.89 & 54.83 & 58.86 \\
BART + MS & 55.13 & 38.49 & 51.92 & 58.55 & 33.52 & 53.77 & 62.83 & 61.67 & \textbf{62.63} & 13.64 & 13.48 & 13.60 & 55.07 & 69.46 & 47.00 & 58.23 \\
% BART + MS + I & 56.94 & 63.99 & \textbf{61.07} & 62.79 & 40.00 & \textbf{57.85} & 61.93 & 34.44 & 59.19 & 13.79 & 14.16 & \textbf{13.88} & \textbf{57.17} & 74.84 & 45.83 & \textbf{60.34} \\
{BART + MS + I} & 56.94 & 36.21 & \textbf{52.96} & 62.79 & 28.61 & \textbf{56.29} & 61.93 & 62.66 & 62.04 & 13.79 & 14.16 & \textbf{13.88} & \textbf{57.17} & 74.84 & 45.83 & \textbf{60.34} \\
\bottomrule
\end{tabular}}
\caption{
Production F1 on the full tree (PF1), top layer (PF1$_I$) and lower
layer (PF1$_L$); exact match accuracy on full tree (EMA$_T$); F1 on
intent classification; F1 and AAA on binary classification. All scores
are computed on the test set and reported as percentages of hateful
(H) and non-hatefull (NH) instances, and their combination (All). Best
performing models are boldfaced.
\label{tab:experiment}}
\end{table*}%

\begin{table*}[t]
\centering
\resizebox{\textwidth}{!}{%
\begin{tabular}{ll}
\toprule
%\textbf{Parsed Post} & \textbf{G} & \textbf{P}\\
%\midrule
1. & Never turn your back on [a brown skinned mud monkey]$_{\text{DehumanComp}}$ They do whatever the [jews]$_{\text{Target, PC}}$ tell them to do\\
2. & I d keep my distance if [I]$_{\text{Target}}$ was a l3zv0z because we re coming for you Strong and hard Beware\\
3.& Yes we gotta keep those [foxes]$_{\text{Target}}$ out [they steal natural resources]$_{\text{DehumanComp}}$ like food at the expense of our pets\\
4. & More [refugees]$_{\text{Target}}$ Cool more the merrier There is literally no reason not to do this\\
5. & \multirow{2}{*}{\shortstack[l]{Children are better off not being adopted and raised by sexually active [couples $\rightarrow$ \texttt{women}]$_{\text{Target, PC}}$ in general [their sexual\\degeneracy will mess]$_{\text{DerogatoryOpinion}}$ up the children they raise but it won t show immediately}} \\
& \\
6. & Yes we gotta keep those [foxes $\rightarrow$ \texttt{apes}]$_{\text{Target}}$ out [they steal natural resources]$_{\text{DehumanComp}}$ like food at the expense of our pets\\
\bottomrule
\end{tabular}}
\caption{
Posts that are incorrectly parsed (but not necessarily incorrectly classified) by our model.
\label{tab:error-analysis}}
\end{table*}%

% \paragraph{Explainability at a small price} TODO.
Our results are summarized in Table~\ref{tab:experiment} (scores are
means over five runs; hyperparameter values can be found in our code
documentation). Our model achieves a production F1 of $52.96\%$,
outperforming all comparison models. When looking at the top level of
the tree (PF1$_{I}$), model performance on hateful instances (H) is
considerably inferior to non-hateful ones (NH).  This is not
surprising, since hateful instances can be represented with
${\sim}{4}$ sketches while non-hateful ones are noiser and can present
a larger number of slot combinations.  Model performance at filling
correctly detected slots for hateful and non-hateful instances is
comparable ($61.93$\% and $62.66$\%), approaching the human
ceiling. EMA$_T$ scores are slightly higher for the non-hateful class,
but this is not unexpected since hateful trees all have at least three
slots, while many non-hateful ones have only one (i.e., a target). 

Our model achieves an F1 of 57.17\% on intent classification. In the
binary setting, F1 jumps to 74.84\%, suggesting that some mistakes on
intent classification are due to the model confusing different hateful
intents. As with all other models in the literature, the AAA-score is
just below random guessing \cite{2021-calabrese-aaa}. Overall,
improvement with respect to baselines is significant for all
metrics. We also observe that both sketches and our intent-aware loss
have a large impact on the quality of the generated trees, and the
intent predictions based on them. PF1$_L$ scores for \mbox{BART + MS} are
higher but these are computed on \textit{correctly} detected slots;
the proportion of correct slots detected by this model is worse than
the full model (see PF1$_I$ for BART+MS vs. BART+MS+I).

%number of correct slots detected by BART + MS is lower than the ones
%detected by our final model.

%each additional component results in a 10-point gain on PF1,
%demonstrating that both the meaning sketches and 
% Hyperparameters?
% \paragraph{Ablation study} TODO. Keep chatting about Table~\ref{tab:experiment}.
% Only Bart
% Bart with meaning sketches AND double decoding
% Bart with meaning sketches and additional classification loss

\subsection{Error Analysis}
We sampled 50 instances from the test set, and manually reviewed the
trees generated by the five variants of our model (one per random
seed). Overall, we observe that error patterns are consistent among
all variants. In posts containing multiple targets, a recurrent
mistake is to link the hateful expression to the wrong target,
especially if the mention of the correct target is implicit (see
example 1 in Table~\ref{tab:error-analysis}).

We also see cases where the parsing is coherent to the selected
target, but this prevents the model from detecting hateful messages
towards a different target (e.g., \textit{``l3zv0z''} in
example~2). {Some mistakes stem from difficulty in distinguishing
  \texttt{DerogatoryOpinion} from other slots, as in example~3 where
  the opinion is misclassified as a dehumanising comparison. This is a
  reasonable mistake, as comparisons to criminals are considered
  dehumanising according to the policy (and therefore annotation
  instructions) and are often annotated as
  \texttt{DehumanisingComparison} in the dataset.} We also observe
that for posts correctly identified as non-hateful, the model tends to
miss out on protected characteristics even when they occur (example
4).  The model also hallucinates values for slots due to stereotypes
prominent in the dataset.  In example 5, \textit{``women''} is
mistakenly generated as the target of a sentence about sexual
promiscuity (of couples), and in example 6 the model hallucinates
\textit{``apes''} as the animal in the comparison.  In future work,
hallucinations could be addressed by explicitly constraining the
decoder to the input post.

Finally, we analysed the behaviour of the model in AAA scenarios, and
observed that it struggles with counter speech, as the negative stance
is often expressed with a negative opinion about the proponent of the
hateful opinion, and therefore tagged as
\texttt{DerogatoryOpinion}. Adding words that correlate
  with the hateful class to non-hateful posts succeeds in
  misleading our model; non-hateful instances often differ from
  hateful ones by a slot, rendering  distractors more
  effective. However, for the same reason, the
  addition of such words can also flip the label (e.g., adding
  \textit{``\#kill''} to a post containing a target and a protected
  characteristic), and the model is incorrectly penalised by AAA
  (which assumes the label remains the same).
% Other limitation is that we're annotating only one possible parsing? So maybe the model is penalised too much
% In general it's easier for the moderator to decide whether to agree or not

%%% Local Variables: 
%%% mode: latex
%%% TeX-master:  "../tacl2021v1-template"
%%% End:

\section{Discussion}

The overwhelming majority of approaches to detecting abusive language
online are based on training supervised classifiers with labelled
examples. Classifiers are expected to learn what abuse is based on
these examples alone. We depart from this approach, reformulate the
problem as policy-aware abuse detection and model the policy
explicitly as an Intent Classification and Slot Filling task. Our
experiments show that conventional black-box classifiers learn to
model \textit{one} of the phenomena represented in the dataset, but
small changes such as different random initialisation can lead the
very same model to learn different ones.  Our \mbox{ICSF-based}
approach guides the model towards learning policy-relevant phenomena,
and this can be demonstrated by the explainable predictions it
produces.

We acknowledge that policies for hate speech, as most human developed
guidelines, leave some room for subjective interpretation. For
instance, moderators might disagree on whether a certain expression
represents a dehumanising comparison. However, the more detailed the
policy is (e.g.,~by listing all possible types of comparisons), the
less freedom moderators will have to make subjective judgments. The
purpose of policies is to make decisions as objective as possible, and 
 our new problem formulation shares the same goal.

 While our model still makes errors, the proposed formulation allows
 us to precisely pinpoint where these errors occur and design
 appropriate mitigation strategies. This is in stark contrast with
 existing approaches, where instability is the consequence of spurious
 correlations in the data, it is hard to isolate errors and,
 consequently, mitigation strategies are often not grounded in human
 knowledge about abuse.
%The error analysis shows that they are more rational than the errors that black-box models make.
%Error analysis shows that most errors result from the expressitivity-complexity trade-off in our \textit{implementation} of the ICSF representation, as slots are all children of the intent node and 
 For example, our analysis showed that our model can sometimes fail to
 generate the correct tree by mixing the targets and sentiments of
 multiple opinions. This suggests that it would be useful to have
 nested slots, e.g., a derogatory opinion as the child of its
 corresponding target.  This could also help the model learn the
 difference between derogatory opinions (nested within a target node)
 and negative stance (nested within an opinion node), facilitating the
 detection of counter speech examples.
%The error analysis further showed that the  model, like all abuse detection models, struggles to recognise counter speech; this suggests that i
 Introducing a slot for the proponent of an opinion could also help,
 as the model would then recognise when a hateful
 opinion is expresed by someone other than the author.
%Main takeaways from our experiments are that 1) our architecture can produce fully explainable predictions without loss in performance, and 2) all models are still far from being robust. However, error analysis shows the errors made by our model to be more ``rational'', as a consequence of a more precise problem formulation and a more thought out architecture design. Most errors are in fact a consequence of the expressitivity-complexity trade-off in our \textit{implementation} of the ICSF representation, as slots are all children of the intent node and the model can fail in generating the right tree by mixing up the targets and sentiments of multiple opinions. 

%Finally, we would also like to emphasize that our modeling approach is
% not policy-specific and could be adapted to other policies used in
% industry or academia and representative datasets
% (e.g., \citealt{2021-vidgen-cad}). Our formulation of the abuse detection
% problem and the resulting annotation are modular and can be modified
% easily, e.g., by removing or adding  intents and slots.
 Finally, we would like to emphasize that our modeling approach is not
 policy-specific and could be adapted to other policies used in
 industry or academia. Our formulation of abuse detection and the
 resulting annotation are compatible with more than one dataset (e.g.,
 \citet{2021-vidgen-cad}) and could be easily modified, e.g.,~by
 removing or adding intents and slots. Extending our approach to other
 policies would require additional annotation effort, however, this
 would also be the case in the vanilla classification setting if one were
 to use a different inventory of labels.
%   with studies wishing to explore different annotation guidelines in
 %  the ``old'' classification setting.}

 %Although we leave the exploration of these paths as future work, the current stage is per se a big progress with respect to the existing approaches, where instability is consequence of spurious correlations in the datasets, and it is hard to design mitigation strategies (which often happen to be completely black box, or not grounded in human experience of abuse).

%We acknowledge that the better defined problem of policy-aware abuse detection comes at the price of policy-specificity, as applying our approach to a different hate speech policy would require modelling such policy and possibly some data annotation. However, we stress how most policies used in industry or academia are identical or very similar to the one we adopted, and more than one dataset is available where posts are annotated with such policy (e.g., \citet{2021-vidgen-cad}). Furthermore, the modularity of our approach easily enables the addition or removal of intents and slots.
% However this is also a limitation: if you want to extend the policy, you need to annotate more data -> But current policy is also compatible with CAD

%%% Local Variables: 
%%% mode: latex
%%% TeX-master:  "../tacl2021v1-template"
%%% End:

%\section{Ethical Statement}
%\input{content/08_ethical_statement}

\section{Conclusions}
% In this work we changed the rules of abuse detection by introducing the concept of Policy-Aware Abuse Detection. In PAAD, humans that agree on the interpretation of a post also agree on its classification label. In its explainable version, the task requires models to also produce human readable explanations that are specific to the input post. To enable models to reason over the policy, we convert a policy for hate speech into a machine-friendly representation, where each policy rule is a conjuction of multiple relations. We collect an English dataset where posts are tagged with such relations, and design a new architecture by adapting and enhancing ICSF technologies for the PAEAD domain. The result is a model which is totally explainable without any loss in performance with respect to existing black-box models. Our model is more robust than existing architectures and more ``rational'' in its mistakes, leaving clear indication of the path to pursue a more robust detection of abusive content.

In this work we introduced the concept of policy-aware abuse detection
which we argue allows to develop more interpretable models and yields
high-quality annotations to learn from. Humans that agree on the
interpretation of a post also agree on its classification label. Our
new task requires models to produce human-readable explanations that
are specific to the input post. To enable models to reason over the
policy, we formalise the problem of abuse detection as an instance of
ICSF where each policy guideline corresponds to an intent, and is
associated with a specific set of slots. We collect and release an
English dataset where posts are annotated with such slots, and design
a new neural model by adapting and enhancing ICSF architectures to our
domain. The result is a model which is more reliable than existing
approaches, and more ``rational'' in its predictions and mistakes. In
the future, we would like to investigate whether and how the
explanations our model produces influence moderator decisions.
%allowing developers to 
%indication of the path to pursue for more robust detection of abusive
%content.

\section*{Acknowledgements}
We would like to thank the participants of our pilot study: Alberto Testoni, Alessandro Steri, Amr Keleg, Angelo Calabrese, Atli Sigurgeirsson, Christina Ovezeik, Cristina Coppola, Eddie Ungless, Elisa Caneparo, Erika Pan, Francesca De Donno, Gautier Dagan, Giuseppe De Palma, Jamila Moraes, Jie Chi, Julie-Anne Meaney, Kristian Iliev, Laura Forcina, Luigi Berducci, Marco Fumero, Maria Luque Anguita, Mariachiara Sica, Melina Gutiérrez Hansen, Nikita Moghe, Sandrine Chausson, Sara Forcina, Verna Dankers, and Wendy Zheng.
We are also grateful to Eddie Ungless, Matthias Lindemann, Nikita Moghe, and Tom Sherborne for their valuable feedback.
Finally, we thank the action editor, Nitin Madnani, and the anonymous reviewers for their helpful comments.
This work was supported in part by Huawei and the UKRI Centre for Doctoral Training in Natural Language Processing, funded by the UKRI (grant EP/S022481/1) and the University of Edinburgh, School of Informatics. Lapata gratefully acknowledges
the support of the UK Engineering and Physical Sciences Research
Council (grant EP/W002876/1) 
and the European Research Council (award 681760).

\begin{figure}[h!]
    \centering
    \begin{subfigure}[b]{0.15\textwidth}
        \centering
        \includegraphics[height=6.2mm]{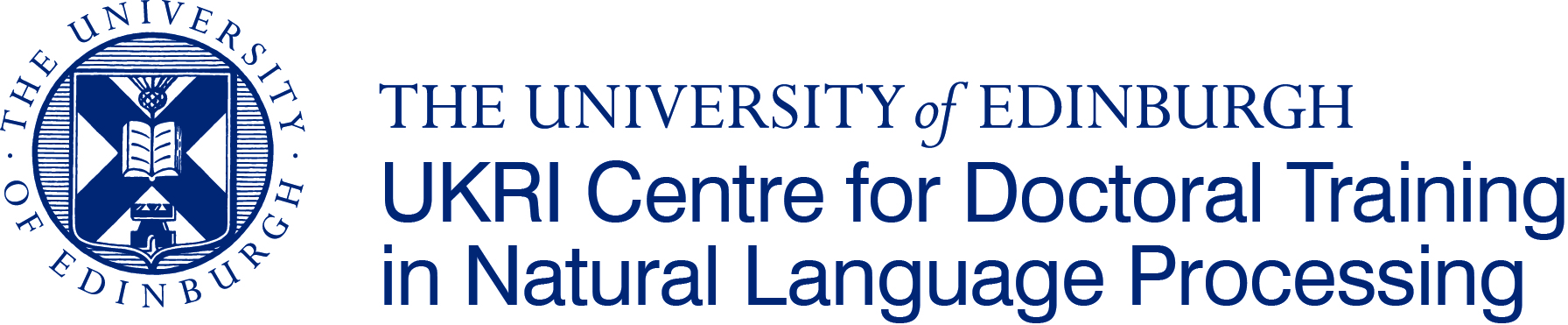}
    \end{subfigure}
    \hfill
    \begin{subfigure}[b]{0.08\textwidth}
        \centering
        \includegraphics[height=5.7mm]{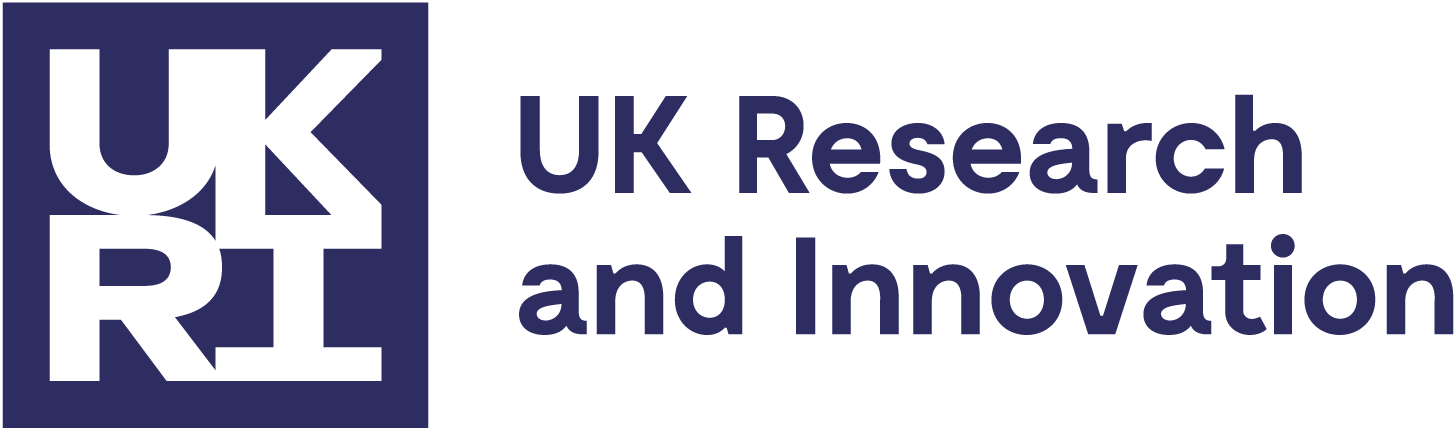}
    \end{subfigure}
    \hfill
    \begin{subfigure}[b]{0.15\textwidth}
        \centering
        \includegraphics[height=6.2mm]{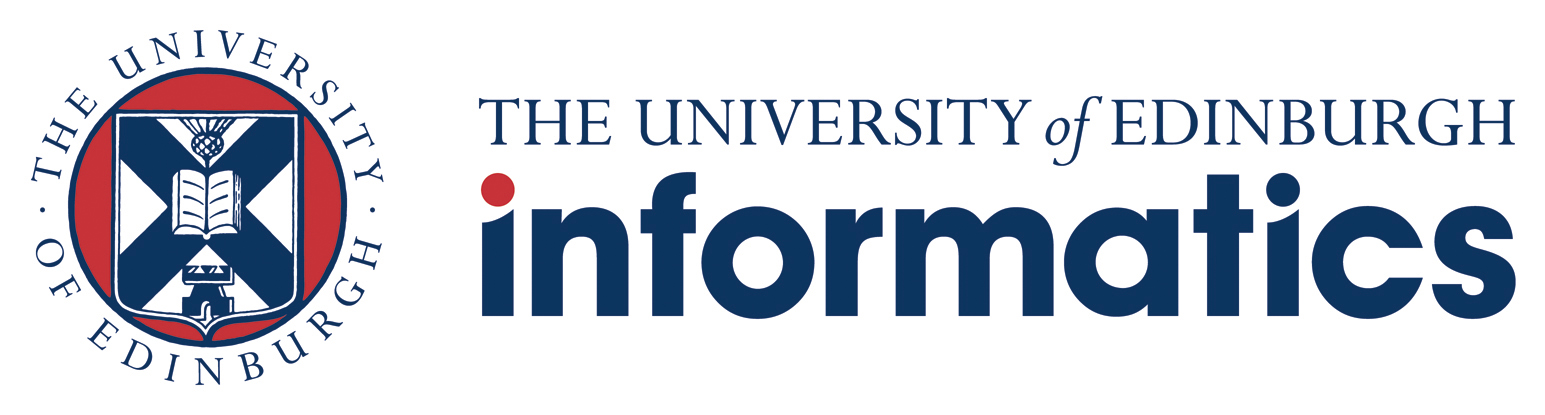}
    \end{subfigure}
\end{figure}

\bibliography{tacl2021}
\bibliographystyle{acl_natbib}

\end{document}